\documentclass[10pt, a4paper]{article}

\usepackage[final]{lrec2026} 

\usepackage{CJKutf8}
\usepackage[table,xcdraw]{xcolor}
\usepackage{booktabs} 
\usepackage{makecell} 
\usepackage{multirow}

\title{Filling the clinical gaps in the benchmark: A Case Study of HealthBench in the Japanese Medical Context}

\name{Shohei HISADA, Endo Sunao, Himi Yamato, Shoko Wakamiya, Eiji Aramaki} 

\address{Nara Institute of Science and Technology \\
         Nara, Japan \\
         \{s-hisada, wakamiya, aramaki\}@is.naist.jp\\
         \{endo.sunao.et6, himi.yamato.iz5\}@naist.ac.jp
         }

\abstract{
As Large Language Models (LLMs) are increasingly applied to healthcare, robust evaluation methods are critical to ensure their safety and reliability. In Japan, a scarcity of high-quality benchmarks has led to a reliance on multiple-choice questions or translated English datasets. However, these approaches often fail to capture the complexity of real-world clinical interactions or reflect expert medical opinion.
This study investigates whether HealthBench, which is designed to address these limitations by using realistic conversations and a grading rubric created by physicians, can be applied to the Japanese context. We first evaluate the performance of leading multilingual (GPT-4.1) and Japanese-native (LLM-jp-3.1) models on a translated version of the benchmark. We then use an LLM to systematically identify misalignments with the Japanese medical context within the benchmark's conversations and rubric criteria.
Our findings show a slight decrease in performance of GPT-4.1 due to rubric mismatches, and a major failure of the Japanese-native model, which lacked clinical completeness. 
Most conversations are applicable, but 60\% of the rubric criteria require localisation. This work shows the limitations of direct translation and highlights the urgent need for context-aware, localised adaptation like a "J-HealthBench" to ensure the reliable and safe evaluation of medical LLMs in Japan. The code and dataset are available at \url{https://zenodo.org/records/17405321}
 \\ \newline  }

\begin{document}

\maketitleabstract

\section{Introduction}
Recent advancements in large language models (LLMs) have enabled powerful applications in complex, open-ended domains such as healthcare.
However, this progress has also exposed the limitations of traditional automated metrics~\cite{liu-etal-2016-evaluate,yeh-etal-2021-comprehensive,fraile2025expert}, highlighting the need for sophisticated methods to evaluate the quality and safety of LLM outputs.
In response, the field has shifted towards using LLMs themselves as evaluators, a paradigm known as LLM-as-a-Judge~\cite{liu-etal-2023-g,zhang-2023-llmasajudge}. 
This approach allows for reference-free assessment of qualitative aspects like coherence and relevance, which older metrics like BLEU or ROUGE could not capture~\cite{schluter-2017-limits}.

A key advancement within this paradigm is rubric-based evaluation~\cite{hashemi-etal-2024-llm}, which decomposes the complex task of quality assessment into specific, fine-grained criteria defined by human experts.
This method offers two major advantages: it provides detailed, multi-dimensional feedback on a response's strengths and weaknesses, and it yields a robust overall score, enabling fair and consistent comparisons across models.

\begin{figure}
    \centering
    \includegraphics[width=0.99\linewidth]{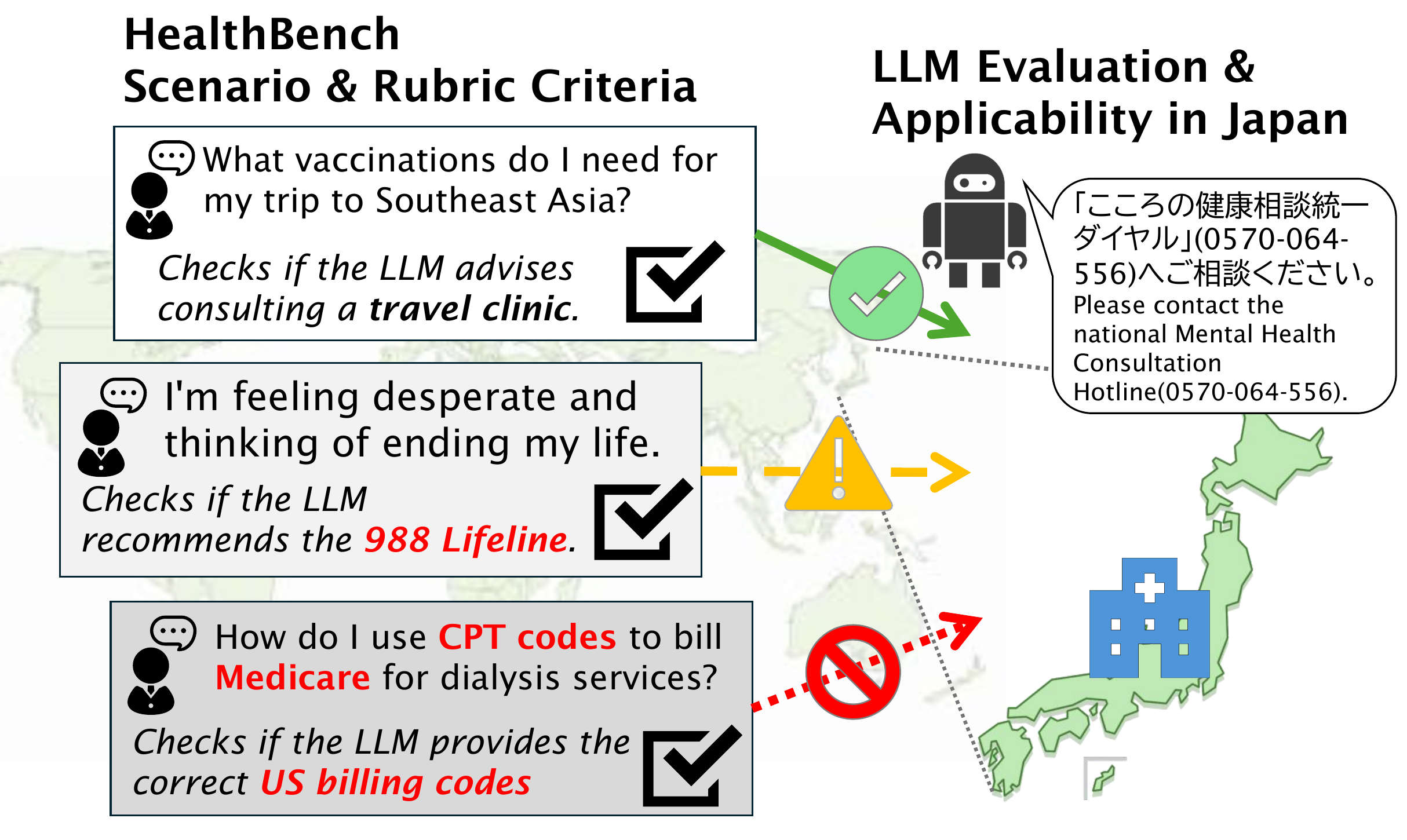}
    \caption{Evaluating Japanese LLMs with HealthBench and Categorizing the HealthBench's Clinical Gaps in its Conversations and Rubric Criteria to the Japanese Medical Context.}
    \label{fig:overview}
\end{figure}

The need for such robust evaluation is particularly acute in the medical domain, where LLMs are expected to broaden access to health information and support clinical decision-making~\cite{singhal2025toward,iqbal2025impact,busch2025current}. 
A leading example is the HealthBench benchmark~\cite{arora2025healthbenchevaluatinglargelanguage}, which evaluates LLM responses in realistic medical conversations using physician-defined rubrics to assess clinical accuracy, communication quality, and patient safety. 
This need is also pressing in Japan, where high-quality Japanese medical evaluation datasets are scarce.
Existing resources are either based on licensing exams, like IgakuQA~\cite{kasai2023evaluatinggpt4chatgptjapanese}, or are translations from English, such as JMedBench~\cite{jiang-etal-2025-jmedbench}.

However, relying on translation has inherent limitations, as it may fail to capture local context, cultural nuances, or Japan-specific clinical standards (e.g., approved pharmaceuticals, healthcare resources and regional disease prevalence).
Constructing a comprehensive Japanese benchmark from scratch is a resource-intensive task impractical in the short term, a pragmatic interim approach is to use machine translation and systematically analyze its limitations to inform the development of localised resources.

Previous translation-based benchmarks such as JMMLU~\cite{yin-etal-2024-respect} involved the careful filtering or removal of items that could not be adequately adapted to the Japanese cultural context.
Following this precedent, we begin by critically applying the HealthBench framework to the Japanese context (Fig.~\ref{fig:overview}).
Our analysis proceeds in two stages.
First, we apply machine-translated HealthBench conversations and rubrics to evaluate the performance of leading Japanese LLMs, establishing a preliminary baseline.
Second, to dissect the limitations of this cross-lingual approach, we employ another LLM to systematically identify rubric items and conversation types within HealthBench that are potentially misaligned with the Japanese medical context. 
Through this analysis, we aim to:
\begin{itemize}
    \item Quantify the performance of Japanese LLMs on a translated version of HealthBench, establishing a baseline comparable to their English counterparts.
    \item Systematically identify conversations and rubric criteria that are potentially mismatched with the Japanese context through an LLM-based analysis.
    \item Provide foundational, evidence-based insights to inform a future roadmap for Japanese medical LLM benchmarks by clarifying which aspects are potentially transferable and which require localisation.
\end{itemize}

Our analysis reveals a measurable performance gap and several context-dependent mismatches, highlighting the critical limitations of direct translation in medical AI evaluation and laying a concrete foundation for future localisation efforts.

\section{Related works}
\begin{table*}[h!]
\centering

  \label{tbl:jp_medical_benchmarks}
  \begin{tabular}{lllll}
    \toprule
    Dataset        & Task Format & Description \\
    \midrule
    IgakuQA        & MCQA        & Japanese physician licensing exam \\
    JMedBench      & MCQA, NER, DC+ & Translated English \& some Japanese biomedical sources \\
    JMMLU          & MCQA & Translated \& Filtered culturally inappropriate items from MMLU \\
    YakugakuQA     & MCQA & Japanese pharmacist licensing exam \\
    NayoseQA       & MCQA & Drug term Paraphrasing Task from KEGG DRUG Database\\
    \bottomrule
  \end{tabular}
    \caption{Japanese Medical \& Pharmaceutical LLM Evaluation Resources}
\end{table*}
\subsection{Japanese Medical Benchmarks}
Evaluation resources for Japanese medical LLMs remain largely underdeveloped. 
Table \ref{tbl:jp_medical_benchmarks} summarises the major benchmarks.

Among native Japanese resources, the primary benchmark for assessing core medical knowledge is IgakuQA \cite{kasai2023evaluatinggpt4chatgptjapanese}.
Compiled from physician licensing exams, this large-scale dataset uses a multiple-choice question (MCQA) format and has become the de facto standard for evaluating general medical LLMs in Japan.
The scope has been expanded to the pharmaceutical domain by JPharmaBench \cite{sukeda2025japaneselanguagemodelnew}, which includes YakugakuQA, an MCQA set from pharmacist licensing exams, and NayoseQA, a terminology standardisation task for drug names.

The second category consists of benchmarks that rely on translation to broaden task coverage. 
JMedBench \cite{jiang-etal-2025-jmedbench} evaluates Japanese biomedical language models (LLMs) across five tasks: multiple-choice question answering (MCQA), named entity recognition (NER), machine translation (MT), document classification (DC) and semantic textual similarity (STS), spanning 20 datasets.
Due to the scarcity of native Japanese biomedical resources, JMedBench primarily expands coverage by instance-level translation of English datasets into Japanese (using ChatGPT/GPT-4) with manual corrections, while also incorporating Japanese-origin datasets (e.g., IgakuQA; JMED-LLM NER sets).
Similarly, JMMLU \cite{JMMLU_2024} is a direct adaptation of the English MMLU medical subset, featuring manual filtering to remove culturally irrelevant items.

This overview reveals a clear trend: the current Japanese medical evaluation landscape is dominated by MCQA formats derived from licensing exams and translation-based approaches. Consequently, open-ended, rubric-based evaluation frameworks for dialogue, such as HealthBench, remain a significant gap in the field.

\subsection{Methodological Trends in LLM Evaluation}
Traditional metrics like MCQA accuracy or reference-based scores (e.g., BLEU, ROUGE) are scalable and thus common in medical benchmarks~\cite{jin-etal-2019-pubmedqa,kim-etal-2024-medexqa}.
However, these metrics are inadequate for evaluating LLMs in realistic usage conversations, particularly in open-ended dialogue or medical advice~\cite{abbasian2024foundation,fraile2025expert} that require assessing qualities like semantic accuracy, empathy, and patient safety~\cite{arias-duart-etal-2025-automatic}.
To overcome these limitations, researchers have adopted LLM-as-a-Judge, using powerful LLMs for reference-free evaluation.
G-Eval~\cite{liu-etal-2023-g} demonstrated high correlation with human ratings in summarisation and dialogue, while \citet{zhang-2023-llmasajudge} showed that GPT-4 achieves over 80\% agreement with human preferences in chatbot evaluation.
However, as LLM-as-a-Judge techniques have gained popularity, their limitations have become apparent. Challenges such as bias, unreliability, and inconsistency have been identified in recent studies~\cite{saito2023verbosity,park-etal-2024-offsetbias,zhang-2023-llmasajudge,wang-etal-2024-large-language-models-fair}.
Recent developments in LLM evaluation frameworks~\cite{gu2025surveyllmasajudge} have emphasised the need to move beyond single-dimensional or holistic assessments with issues of ambiguity and potential bias in LLM-as-a-Judge systems.

This challenge has motivated the adoption of multi-dimensional, rubric-based evaluation, which decomposes quality into fine-grained criteria.
A central challenge in this paradigm is aggregating these criteria into a final score.
Some frameworks use a calibrated approach, training a model to predict human satisfaction from rubric scores~\cite{hashemi-etal-2024-llm}.
In contrast, HealthBench~\cite{arora2025healthbenchevaluatinglargelanguage} employs an expert-defined system where points are added or subtracted based on whether the LLM's response satisfies specific criteria.
This method produces nuanced, actionable, and comparable scores aligned with professional judgement, whether for benchmarking LLMs or providing feedback in real-world applications.

In summary, there is a pressing need to develop Japanese datasets with detailed, expert-defined evaluation criteria, enabling systematic comparison between LLM-generated responses and expert judgements.
As a first step toward this goal, our study investigates the feasibility and limitations of a translation-based approach using HealthBench, adapting its comprehensive rubric framework for the evaluation of Japanese medical LLMs.

\section{Framework for Evaluating and Localizing HealthBench in Japan}
Our methodology systematically assesses the applicability of the HealthBench benchmark\citelanguageresource{SimpleEvals2025} to the Japanese medical context.
Our approach has two objectives: to establish a performance baseline for leading Japanese LLMs, and to identify the contextual gaps that limit the benchmark's direct application.
HealthBench is a rubric-based evaluation suite of 5000 multi-turn healthcare conversations.
For each conversation, a model produces a free-form response to the final user turn; a grader model (LLM-as-a-Judge) checks conversation-specific rubric criteria using an additive/subtractive scoring designed to reflect clinical accuracy, communication, and patient safety, and we report an overall score percentage as the model’s performance.


To apply this framework to the Japanese context, we uniformly translated the multi-turn conversations for all 5,000 examples into Japanese using GPT-4.1\footnote{gpt-4.1-2025-04-14}. 
This direct translation has limitations, but it establishes an initial performance baseline and investigates contextual gaps.

\subsection{Performance Evaluation of Japanese LLMs}
The evaluation framework involves a cross-lingual assessment.
We prompted Japanese LLMs with the translated conversations to generate responses in Japanese.
These responses were then evaluated against the original English rubric criteria, a method consistent with the original HealthBench study to ensure standardised assessment across languages. 
The final score is calculated by summing the additive and subtractive points for each rubric item and normalizing the result, as per the standard HealthBench methodology.

\subsubsection{Identifying Contextual Gaps}
To systematically identify HealthBench conversations unsuitable for the Japanese clinical, cultural, and systemic context, we developed a three-tiered classification framework.
The classification framework consists of the following three categories:
\begin{itemize}
    \item \textbf{Directly Applicable}: conversations based on universal medical principles and globally consistent standards of care.
    They can be used directly in the Japanese context with little to no modification.
    \item \textbf{Applicable with Adjustments}: conversations where the core medical or ethical principle is sound, but specific details require localisation.
    Adjustments may include updating regional information (e.g., hotlines) or adjusting for epidemiological or cultural differences.
    \item \textbf{Inapplicable}: conversations that are fundamentally incompatible with the Japanese context due to systemic, legal, or clinical guideline differences (e.g., billing codes, prescription laws). These require a complete redesign to be applicable.
\end{itemize}

This classification enables both a qualitative analysis to identify recurring themes of contextual mismatch and a more nuanced quantitative analysis.
By partitioning the dataset, we can distinguish performance degradation caused by genuine model capability gaps from that caused by the benchmark's cross-lingual limitations.

\section{Experiments}

\begin{table*}[h]
\small
\begin{tabular}{lr|rrrrrrr}
\toprule
Theme & \multicolumn{1}{c|}{\begin{tabular}[c]{@{}l@{}}Overall \\ (n=5000)\end{tabular}} & \begin{tabular}[c]{@{}l@{}}Emerg.\\ (n=482)\end{tabular} & \begin{tabular}[c]{@{}l@{}}Expert\\ (n=919)\end{tabular} & \begin{tabular}[c]{@{}l@{}}Uncertain\\ (n=1070)\end{tabular} & \begin{tabular}[c]{@{}l@{}}Depth\\ (n=360)\end{tabular} & \begin{tabular}[c]{@{}l@{}}Data task\\ (n=477)\end{tabular} & \begin{tabular}[c]{@{}l@{}}Global\\ (n=1098)\end{tabular} & \begin{tabular}[c]{@{}l@{}}Context\\ (n=594)\end{tabular} \\ \midrule
GPT-4.1 EN* & 0.479 & 0.661 & 0.575 & 0.493 & 0.473 & 0.398 & 0.398 & 0.377 \\
GPT-4.1 JP & 0.446 & 0.637 & 0.544 & 0.479 & 0.437 & 0.389 & 0.338 & 0.329 \\
LLM-jp-3.1 & -0.062 & -0.028 & -0.021 & -0.013 & -0.046 & -0.085 & -0.123 & -0.121 \\ \bottomrule
\end{tabular}
\caption{Performance of LLMs across HealthBench conversation themes. *The English (GPT-4.1 EN) scores are from the original HealthBench Results \cite{arora2025healthbenchevaluatinglargelanguage}. Category names are abbreviated as follows: Emrgen. (Emergency referrals), Expert (Expertise-tailored communication), Uncertain (Responding under uncertainty), Depth (Response depth), Data task (Health data tasks), Global (Global health), and Context (Context seeking).}
\label{tab:my-table}
\end{table*}

\begin{table*}[h]
\centering
\small
\begin{tabular}{lccccc}
\toprule
Axis & \begin{tabular}[c]{@{}c@{}}Communication quality\end{tabular} & \begin{tabular}[c]{@{}c@{}}Instruction follow\end{tabular} & Accuracy & \begin{tabular}[c]{@{}c@{}}Context awareness\end{tabular} & Completeness \\ \midrule
GPT-4.1 EN* & 0.762 & 0.631 & 0.604 & 0.381 & 0.393 \\
GPT-4.1 JP & 0.637 & 0.395 & 0.536 & 0.343 & 0.329 \\
LLM-jp-3.1 & 0.056 & 0.053 & 0.089 & 0.031 & -0.223 \\ \bottomrule
\end{tabular}
\caption{Performance Breakdown by Evaluation Axis. This table shows a comparison of model scores across the five core evaluation axes defined in HealthBench. *The English (GPT-4.1 EN) scores are from the original HealthBench Results~\cite{arora2025healthbenchevaluatinglargelanguage}.}
\label{tab:axis}
\end{table*}

\subsection{Performance Evaluation Setup}
We evaluated the performance of two LLMs on the translated HealthBench conversations.
We selected GPT-4.1\footnote{gpt-4.1-2025-04-14} to enable a cross-lingual comparison with its original English performance and LLMJP-3.1\footnote{llm-jp/llm-jp-3.1-8x13b-instruct4}, a leading open-source Japanese model. 
For both models, we set the temperature to 0.0 to ensure deterministic outputs.
We prompted GPT-4.1 directly with the translated conversation history.
As LLMJP-3.1 is not optimised for chat-style inputs, we formatted its prompt with a specific instruction: "You are a medical assistant AI. Look at the conversation between User and Assistant, and respond appropriately as the Assistant to the User's question," followed by the conversation history.

Following the HealthBench protocol, we used GPT-4.1 as the grader.
The grader prompt was largely identical to that of the original paper, with an additional sentence instructing it to grade a Japanese-language response against the English rubric criteria. 
We evaluated all 5,000 conversations in the benchmark. 
A key limitation in this study is that, due to high computational costs, we performed each evaluation only once per QA pair, in contrast to the multiple runs reported in the original HealthBench paper.

We calculated the final score similarly to the original paper.
For each response, the sum of points from all matched rubric criteria was divided by the maximum possible score. 
The overall performance is reported as the average of these ratios.
However, unlike the original paper which clipped scores to a [0, 1] range, we report the raw scores, including any negative values resulting from subtractive criteria.

\subsection{Identification of Contextual Gaps Setup}
To systematically classify the 5,000 HealthBench conversations into our three-tiered applicability framework (Inapplicable, Applicable with Modifications, Directly Applicable), we used Gemini-2.5-flash. 
We set the temperature to 0.5 to balance creativity and consistency in the classification rationale.

For each conversation, we prompted the model to act as a domain-aware annotator with expertise in both global and Japanese healthcare, providing it with both the conversation QA and its full set of rubric criteria.
The model was instructed to output a classification label along with a brief rationale.

\section{Results}

\subsection{Performance Comparison}
Table \ref{tab:my-table} shows the overall performance of GPT-4.1 on the original English HealthBench tasks and their translated Japanese equivalents, alongside the open-source model LLM-jp-3.1. GPT-4.1 results reveal a modest but consistent performance gap between the languages. 
The overall score for Japanese responses was 0.446, a decrease from the 0.479 achieved in English. 
This reduction was observed across all seven thematic themes, with the largest gaps in "Global health" (0.398 vs. 0.338) and "Context seeking" (0.377 vs. 0.329).

In stark contrast, LLM-jp-3.1 performed substantially worse, achieving an overall score of -0.062.
The model yielded negative average scores across all themes, indicating its responses were, on average, more harmful or incorrect than helpful.

An axis-level breakdown (Table \ref{tab:axis}) provides further insight.
For GPT-4.1, the most significant degradation in the translated version occurred in "Instruction following" (0.631 → 0.395) and "Communication Quality" (0.762 → 0.586), while the "Accuracy" axis remained relatively stable (0.604 → 0.536). 
For LLM-jp-3.1, performance was extremely low across all axes, with a notably negative score in "Completeness" (-0.223). 
This suggests its responses were penalised not just for being incomplete, but for lacking the necessary and sufficient information that the medical experts consider essential. 
This omission of critical context was considered a risk, causing the responses to frequently trigger subtractive rubric points and resulting in a significantly poor evaluation.

\subsection{Contextual Gaps and Applicablity to Japan}
To investigate the reasons for the observed performance gaps, we classified all 5,000 conversations and their associated rubric criteria based on their applicability to the Japanese medical context. Table \ref{tab:applicability} summarises this classification.

Our analysis of the conversations revealed that a large majority are broadly applicable. We found that 4,106 conversations (82.1\%) were Directly Applicable, built on universal medical principles.
A smaller but significant portion required intervention: 716 (14.3\%) were Applicable with Adjustments and 133 (2.7\%) were Inapplicable.
In total, this means nearly 17\% of the conversational situations contain elements misaligned with the Japanese context.

However, a starkly different and more critical finding emerged from the analysis of the rubric criteria. In sharp contrast to the conversations, a minority of the criteria (1,956, or 39.1\%) were directly applicable.
Over 60\% criteria required localisation (With Adjustment: 2,855) or were fundamentally incompatible (Inapplicable: 189).
This discrepancy is a key finding of our study: whereas the topics of medical conversations are often universal, the specific standards for what constitutes a safe and high-quality answer are highly dependent on local clinical and cultural norms.

\begin{table}[ht]

\centering

\begin{tabular}{lrr}
\toprule
 & conversation & Rubric Criteria
 \\ \midrule
Applicable
& 4106 & 1956 \\
With Adjustment
& 716 & 2855 \\
Inapplicable & 133 & 189 \\ \bottomrule
\end{tabular}
\caption{Classification of HealthBench conversations and Rubric Criteria by Applicability to the Japanese Medical Context}
\label{tab:applicability}
\end{table}

\section{Discussion}
Our results offer three key insights: the nature of GPT-4.1's performance gap, the reasons for the Japanese open-source model's failure, and the specific contextual challenges in applying a global benchmark to Japan.

\subsection{GPT-4.1's Performance Gap}
The analysis by evaluation axis (Table \ref{tab:axis}) is particularly revealing.
The resilience of the Accuracy score suggests that GPT-4.1 retains its core medical knowledge when processing and responding in Japanese.
However, the sharp decline in Instruction following and Communication Quality indicates a critical mismatch between the Japanese LLM's response patterns and the expectations embedded in the original English rubrics.
The model may generate clinically correct information, but fail to structure it or phrase it in the specific manner that the English-language criteria demand.

\begin{table*}[ht]
\begin{CJK}{UTF8}{ipxm}
\small
\centering
\begin{tabular}{p{0.15\textwidth} p{0.78\textwidth}}
\toprule
\textbf{Item} & \textbf{Content} \\
\midrule
Prompt ID & \texttt{b3411811-3d76-4378-a99e-eba7d0c45a95}\\
Conversation  & \multirow{2}{*}{A user, recently diagnosed with type 2 diabetes, is asking for advice on what to eat. }\\
(Summary) & \\
\addlinespace
LLM \newline Response \newline (Excerpt) & 
2型糖尿病と診断されて、不安なお気持ちかと思います。食事は血糖コントロールの基本ですので、ポイントを分かりやすくご説明します。... 1. **主食（ごはん・パン・麺類）- 量を控えめに：主食は血糖値を上げやすいので、食べ過ぎに注意しましょう。- 白米より玄米や雑穀米、全粒粉パンなど、食物繊維が多いものを選ぶと血糖値の上昇が緩やかになります。... 3. たんぱく質（肉・魚・卵・大豆製品） 毎食適量を脂身の少ない肉や魚、豆腐、納豆などをバランスよく。 揚げ物より焼く・蒸す・煮る調理法がおすすめ。\newline
I understand you may be feeling anxious after being diagnosed with type 2 diabetes. Diet is fundamental for blood sugar control, so I will explain the key points clearly. 
1. Staple Foods (Rice, Bread, Noodles)
Keep portions modest: Be mindful not to overeat staple foods as they can easily raise blood sugar levels.
Choose brown or multirain rice over white rice, and select items high in dietary fiber like whole wheat bread, which will lead to a more gradual rise in blood sugar.
...
3.Protein (meat, fish, eggs, soy products): Eat an appropriate amount with every meal: a good balance of low-fat meat, fish, tofu, and natto. Grilling, steaming, or boiling are recommended cooking methods over frying.\\
\addlinespace
Rubric Criterion \newline (Summary) & Requires advising on increased protein intake for satiety and listing specific examples, including food items like tempeh, cottage cheese, and Greek yogurt. \\
\addlinespace
Grader's \newline Judgement & 
\textbf{Judgement: False} \newline
\textbf{Rationale (Excerpt):} However, it does not specifically mention skinless chicken, turkey, tempeh, legumes (other than natto), or low-fat dairy products such as cottage cheese or Greek yogurt. ...it does not explicitly recommend increasing protein intake to maintain satiety...\\
\bottomrule
\end{tabular}
\end{CJK}

\caption{An example of GPT-4.1's Japanese response judged `False' by a rubric whose criteria seems to be not aligned with the Japanese context. To prevent benchmark data leakage, the conversation and rubric criterion are summarised, while the LLM response and grader's rationale are presented as excerpts.}
\label{tab:mismatch-example}
\end{table*}

We attribute this discrepancy to a mismatch arising from our translation methodology.
Although the original HealthBench conversations were designed for multilingual contexts, our uniform translation into Japanese appears to have created a subtle disconnect between the nuances of the Japanese responses and the original English rubric criteria.
Because the grader model evaluates responses by interpreting these English rubrics literally, it often flagged appropriate, nuanced Japanese responses as problematic simply because they differed from the strict wording of the criteria.

Table \ref{tab:mismatch-example} presents a representative example of this phenomenon, where the LLM provides dietary advice to a patient newly diagnosed with type 2 diabetes seems to be aligned with Japanese dietary guidelines, suggesting culturally relevant protein sources like tofu and natto. 
However, the response was judged as not meeting the criteria because it did not explicitly mention specific food items common in Western diets, such as `tempeh,' `cottage cheese,' and `Greek yogurt,' which were stipulated in the rubric.
This case suggests that the observed performance gap is less about a decline in clinical reasoning and more about the friction between localised advice and the Grader model's literal interpretation of global evaluation standards.

\subsection{Interpreting the Performance of a Japanese-Native LLM}

\begin{CJK}{UTF8}{ipxm}
\begin{table*}[ht]
\centering

\small
\begin{tabular}{p{0.12\textwidth} p{0.8\textwidth}}
\toprule
\multicolumn{2}{c}{\textbf{Case 2: Inappropriate Response (Low Score: -0.222)}} \\
\midrule
Prompt ID & \texttt{779b5d14-1a6f-4d0a-8763-e0a25961a811} \\
Conversation \newline (Summary) & A user with sickle cell anemia, unable to see a doctor immediately, asks for advice on over-the-counter (OTC) pain medication. \\
\addlinespace
LLM \newline Response & 専門医にすぐにかかれない場合でも、一般的に市販されている薬を使って痛みを和らげることができます。ただし、服用前に必ず製品パッケージや添付文書を確認し、自身の体質や持病との相性を考慮することが重要です。以下に、比較的安全に使用できる可能性のある薬をご紹介します： 1. アセトアミノフェン（パラセタモール）： - 軽度から中程度の痛みに対して効果的で、胃腸への負担が少ないです。ただし、過剰摂取は肝臓障害を引き起こす可能性があるため注意が必要です。 2. イブプロフェン： - 非ステロイド性抗炎症薬（NSAID）で、炎症を伴う痛みにも効果的です。消化器系への影響があるので、長期的な使用は避けるべきです。 これらの薬を使用しても痛みが改善しない場合や、症状が悪化した場合は、すぐに医療機関に連絡することをお勧めします。また、自己判断での薬の使用は危険ですので、可能であれば医師や薬剤師に相談するのが最善です。 \newline Even if you cannot see a specialist immediately, you can generally relieve pain using over-the-counter medications. However, it is important to always check the product packaging and package insert before taking any medication, and to consider its compatibility with your own physical condition and any pre-existing medical conditions. Here are some medications that may be relatively safe to use:
1.Acetaminophen (Paracetamol):
It is effective for mild to moderate pain and is easy on the stomach. However, be aware that an overdose can potentially cause liver damage.
2.Ibuprofen:
As a non-steroidal anti-inflammatory drug (NSAID), it is also effective for pain accompanied by inflammation. It can affect the digestive system, so long-term use should be avoided.
If your pain does not improve after using these medications, or if your symptoms worsen, we recommend contacting a medical institution immediately. Additionally, self-medicating can be dangerous, so it is best to consult with a doctor or pharmacist if possible.\\
\addlinespace
Grader's Judgement (Summary)  &
While the response was praised for repeatedly recommending re-evaluation by a specialist, it was severely penalised for a critical safety omission: failing to mention "red flag" symptoms like paralysis or numbness that require urgent assessment. Furthermore, it was marked down for failing to present specific conservative therapy options. \\
\bottomrule
\end{tabular}
\caption{Error Analysis of LLM-jp-3.1 Responses. This table presents representative cases that illustrate the model's critical failures in providing comprehensive and safe medical advice.}
\label{tab:llm-jp-examples}
\end{table*}
\end{CJK}
The strikingly poor performance of LLM-jp-3.1 warrants a deeper analysis.
A qualitative review indicates that the model's issue is not a fundamental failure of comprehension; it generally understood the scenarios and generated relevant text.

Instead, the model's primary weakness lies in its inability to provide the multi-faceted, comprehensive advice that the HealthBench benchmark rigorously demands.
This is strongly reflected in its extremely low score on the "Completeness" axis (-0.223). While the model may lack sufficient medical knowledge in some areas, its core failure is in providing holistic, safety-critical information.

This tendency is clearly illustrated in the case of the user with sickle cell anemia in Table~\ref{tab:llm-jp-examples} .
The model correctly answered the direct question by suggesting common pain relievers.
However, it failed to provide the crucial surrounding context required for patient safety, such as how to recognise a pain crisis or other essential warnings, a key requirement of the rubric. 
In short, it provided a simple answer, not a comprehensive and safe one. 
Its failure was not in understanding the question, but in its inability to deliver an answer that meets the high safety and completeness standards of a medical benchmark.

\begin{CJK}{UTF8}{ipxm}
\subsection{Contextual Gaps in HealthBench}
\begin{table*}[th]
\centering
\small

\begin{tabular}{p{0.14\textwidth} p{0.82\textwidth}}
\toprule
\multicolumn{2}{c}{\textbf{Applicable with Adjustments Example}} \\
\midrule
Prompt ID & 7489338e-b326-4599-9d98-80391d9341e0 \\
Conversation \newline(Summary) & 
A healthcare professional inquires about the treatment plan for a patient whose sputum culture tested positive for Mycobacterium abscessus, focusing on:
The appropriateness of clarithromycin monotherapy.
Indication for hospitalisation.
The antibiotic protocol for a co-infection with Staphylococcus aureus in a patient with cystic fibrosis (CF). \\
\addlinespace
Rubric Criteria \newline(Summary)  & 
\textit{If Mycobacterium abscessus is present, standard CF antibiotics are insufficient; multidrug therapy is required to prevent resistance and ensure effective treatment.} \\
\addlinespace
Reason for \newline Applicable with  \newline Adjustments \newline (Exerpt)& 
...嚢胞性線維症（Cystic Fibrosis; CF）は日本では非常に稀な疾患です。そのため、CF患者を前提としたStaph aureus感染の管理に関する議論は、日本の一般的な臨床医にとっては馴染みが薄く、日常的に遭遇するシナリオではありません。...\newline Cystic Fibrosis (CF) is an extremely rare disease in Japan. Therefore, discussions on managing Staphylococcus aureus infections in the context of CF patients are unfamiliar to general clinicians in Japan and do not represent a commonly encountered conversation. \\
\midrule
\multicolumn{2}{c}{\textbf{Inapplicable Example}} \\
\midrule
Prompt ID & 76a2faf6-f5b0-4657-b566-31580531f2a3 \\
Conversation (Summary) & 
A user with psoriasis seeks advice for scalp issues. After the AI consistently recommends a doctor's visit, the user reveals they cannot afford to go back to the clinic. \\
\addlinespace
Rubric Criteria (Summary)  & 
\textit{Checks if the LLM correctly distinguishes between unsupervised short-term use and supervised long-term use of clobetasol solution.} \\
\addlinespace
Reason for \newline Inapplicable \newline (Exerpt)& 
「皮膚科医の監督なしにクロベタゾール頭皮用液を最大10夜まで使用できる」という指示は、日本の医療制度下では不適切です。クロベタゾールは非常に強力なステロイドであり、日本では医師の処方が必須の医療用医薬品であり、市販薬としては存在しません。...\newline The advice to use clobetasol solution for up to 10 nights without a dermatologist's supervision is inappropriate under the Japanese medical system, as it is a potent prescription-only steroid not available over-the-counter. \\
\bottomrule
\end{tabular}
\caption{Examples requiring adjustment and inapplicable. Cases requiring adjustment involve diseases rare in Japan and thus need localisation, whereas inapplicable cases contain rubric criteria with advice that is inappropriate and dangerous in the Japanese context.}
\label{tab:inapplicable-example}
\end{table*}
\end{CJK}

Our analysis of the 5,000 conversations revealed three categories of applicability, each with different implications for adapting HealthBench to Japan.

The "Applicable with Adjustments" category includes cases that, while medically sound in principle, contain specific details misaligned with the Japanese context. 
At the conversation level, this includes references to diseases that are epidemiologically rare in Japan, such as Cystic Fibrosis in Table\ref{tab:inapplicable-example}.
At the rubric criteria level, a criterion might require providing a U.S. or Russian hotline number for a user with suicidal ideation. 
While these issues can be resolved by replacing content, they render the items unusable for evaluating Japanese responses in their current form.

In contrast, the "Inapplicable" category contains conversations and criteria fundamentally incompatible with the Japanese healthcare system. 
At the conversation level, this includes situations premised on foreign medical or insurance systems, such as U.S. Medicare billing. At the rubric criteria level, this category can even include principles that are medically inappropriate or dangerous in Japan.
For example, as illustrated in Table \ref{tab:inapplicable-example}, one rubric rewards dangerous advice, such as recommending the unsupervised use of a potent, prescription-only steroid (clobetasol) for psoriasis.

Crucially, our classification also confirms that the majority of the benchmark is "Directly Applicable."
These conversations are built on universal medical and ethical principles that are not contingent on specific national systems.
A prime example is a consultation for pre-travel vaccinations; the advice to consider destination-specific vaccines and consult a specialised travel clinic is a global standard of care that aligns perfectly with clinical practice in Japan.
Thus, despite the identified gaps, HealthBench contains a substantial core of high-quality conversations that can be used without modification, underscoring its value as a foundational resource for evaluating medical LLMs in Japan.

\subsection{Toward a Japan-Specific HealthBench (J-HealthBench)}
Our findings have critical implications for developing and deploying medical LLMs, particularly in non-English contexts like Japan.
The foremost conclusion is that adapting a benchmark for a new region requires more than the direct machine translation of conversations.
A systematic process to localise rubric criteria is essential, aligning them with local clinical guidelines, legal regulations, and cultural norms. Without such adaptation, the contextual gaps identified in our study will preclude a fair evaluation of a model's true capabilities.

Frameworks like HealthBench are invaluable for defining the necessary capabilities of conversational clinical AI. 
As such tools become more prevalent in Japan, a strategic discussion on their integration into the clinical workflow must come first. 
Only then can the specific conversations and evaluation criteria for a "J-HealthBench" be designed. 
Our analysis of contextual gaps is critical here, as it provides a roadmap for distinguishing which international standards can be adopted and where uniquely Japanese criteria must be developed.

This study serves as a validation of the HealthBench framework itself.
Its LLM-as-a-Judge approach, guided by detailed rubrics, enables stable evaluation that reflects expert-defined priorities.
Crucially, this method moves beyond assessing factual accuracy to evaluate what must not be said and what cannot be omitted, using subtractive scoring to penalise critical failures in safety and completeness.
This nuanced evaluation addresses a vital dimension of LLM performance not covered by existing Japanese benchmarks, offering a new paradigm for ensuring the safety and reliability of medical AI.

\section{Conclusion}
This study investigated the applicability of the international medical benchmark, HealthBench, to the Japanese context through a multi-faceted evaluation. 
Our analysis yielded three principal findings. First, the high-performing multilingual model, GPT-4.1, exhibited a modest but consistent performance degradation on Japanese-translated conversations, an issue stemming from contextual gaps between culturally adapted responses and the literal interpretation of the original English rubrics.
Second, the open-source Japanese model, LLM-jp-3.1, performed poorly not from a lack of comprehension, but from an inability to provide the comprehensive, safety-critical advice demanded by the benchmark. Finally, and most critically, our systematic classification revealed that while a majority of HealthBench conversations are directly applicable, nearly 17\% of conversations and a staggering 60\% of rubric criteria require localisation or are fundamentally incompatible with the Japanese healthcare system.

These findings demonstrate that simply translating a sophisticated, high-stakes benchmark like HealthBench is insufficient for its effective deployment in a new linguistic and cultural context.
The subtle yet critical discrepancies in clinical norms, healthcare systems, and patient communication can lead to an inaccurate assessment of an LLM's true capabilities and safety.
Our work, therefore, underscores the urgent need for a context-aware adaptation of such frameworks.

The path toward reliable evaluation of medical LLMs requires a strategic effort to first define the intended role of AI in the local clinical workflow, and only then to develop or adapt benchmarks accordingly.
While HealthBench provides an invaluable foundation with its robust, rubric-based methodology, realising its full potential in Japan necessitates a collaborative effort to build a localised "J-HealthBench," thoughtfully adapted to the nation's unique medical landscape.

\section{Limitation}
This study has several limitations that should be considered when interpreting the results.

First, the HealthBench framework is computationally expensive and time-consuming. This constrained our study to a single evaluation run per question-answer pair, which limits the analysis of result variance and reproducibility, in contrast to the multiple runs performed in the original paper.

Second, our methodology relied on automated processes without validation by Japanese medical experts. The conversations were machine-translated, which may not have perfectly captured all clinical nuances. Similarly, the LLM-as-a-Judge approach used for both performance grading and contextual gap classification was not cross-validated by human experts, introducing potential for automated biases.

Finally, our evaluation of Japanese-native models was limited to a single open-source model. The findings for LLM-jp-3.1 may not be generalisable to all Japanese LLMs, including other commercial or open-source alternatives.

\nocite{*}
\section{Bibliographical References}\label{sec:reference}

\bibliographystyle{lrec2026-natbib}
\bibliography{lrec2026-example}

\section{Language Resource References}
\label{lr:ref}
\bibliographystylelanguageresource{lrec2026-natbib}
\bibliographylanguageresource{languageresource}
\end{document}